\documentclass{IOS-Book-Article}

\usepackage{mathptmx}
\usepackage{soul}\setuldepth{article}

\usepackage{algorithm}
\usepackage{algpseudocode}
\usepackage{graphicx}

\def\hb{\hbox to 11.5 cm{}}

\newtheorem{definition}{Definition}

\begin{document}

\pagestyle{headings}
\def\thepage{}
\begin{frontmatter}              

\title{TaBIIC: Taxonomy Building through Iterative and Interactive Clustering}


\author[A]{\fnms{Mathieu} \snm{d'Aquin}\orcid{0000-0001-7276-4702}%
\thanks{Corresponding Author: Mathieu d'Aquin, \email{mathieu.daquin@loria.fr}}},

\runningauthor{M. d'Aquin}
\runningtitle{TaBIIC}
\address[A]{K Team, LORIA CNRS/INRIA/Université de Lorraine, Nancy, France}

\begin{abstract}
Building taxonomies is often a significant part of building an ontology, and many attempts have been made to automate the creation of such taxonomies from relevant data. The idea in such approaches is either that relevant definitions of the intension of concepts can be extracted as patterns in the data (e.g. in formal concept analysis) or that their extension can be built from grouping data objects based on similarity (clustering). In both cases, the process leads to an automatically constructed structure, which can either be too coarse and lacking in definition, or too fined-grained and detailed, therefore requiring to be refined into the desired taxonomy. In this paper, we explore a method that takes inspiration from both approaches in an iterative and interactive process, so that refinement and definition of the concepts in the taxonomy occur at the time of identifying those concepts in the data. We show that this method is applicable on a variety of data sources and leads to taxonomies that can be more directly integrated into ontologies.
\end{abstract}

\end{frontmatter}

\section{Introduction}

Taxonomies are an integral part of ontologies, whether they represent an aspect of highly axiomatized ontologies in OWL~\cite{OWL}, or simpler concept structures such as those constructed using SKOS~\cite{SKOS}. While ontology building methodologies (see, for example,~\cite{methontology,neon,ontoknowledge}) focus on identifying and defining core concepts, understanding how to hierarchically divide such concepts is often not trivial and is not as well covered by those methodologies. In addition, when the aim is to use such a taxonomy to classify entities into the corresponding categories from existing data, it is important to ensure that the definitions of concepts and subconcepts align well with the attributes and groups that exist in the data, as well as with what is meaningful in the domain. 

For these reasons, some of the tools aiming to support ontology building have focused on (semi)automatic ways to construct such taxonomies in a bottom-up manner, that is,  from existing data. We describe such works in more detail in the following section, focusing on approaches taking structured data as input (i.e., ignoring the many approaches building taxonomies from text). In summary, however, depending on the specific technique applied, those approaches suffer from a number of drawbacks that significantly limit their applicability. Some, for example, are constrained to using only nominal data as input and come up with very large and complex hierarchies that require to be pruned. Others rely on grouping data objects together, coming up with flat structures, or hierarchies and definitions that are difficult to reconnect with intensional definitions of concepts. 

Taking inspiration from these techniques, we explore here the use of an iterative and interactive method for building taxonomies from data. The idea is to enable the user to edit and label subconcepts in the taxonomy as they are being created, therefore, ensuring that what is created is aligned with both the domain and the data. We describe the principles and implementation of a tool called TaBIIC (Taxonomy Building through Interactive and Iterative Clustering) and show how it applies to a variety of data inputs to create taxonomies that match the user's view of the domain, are ready (compatible in format and structure) for integration in ontologies, and are aligned with the source data. \\


In the following (Section~\ref{sec:rw}) we consider common approaches to building taxonomies from structured data. The form of structured data and of the taxonomies created can vary from one approach to the other, even if similarities exist. Therefore, we first introduce some notation and elements of vocabulary in relation to datasets and concept taxonomies (inspired by formal concept analysis and description logics) that can be used as a shared framework to describe existing work. 

\section{Preliminaries}

As mentioned above, we consider the semiautomatic construction of taxonomies from structured data. Although structured data can come in a wide variety of shapes and formats, we restrict ourselves to a common and widely available one: data tables or data frames. 

\begin{definition}[Dataset]
A dataset $D=(O,A)$ is defined as a set of objects $O$ (lines), described through a list of attributes $A$ (columns). Each object $o \in O$ is described as a vector of size $|A|$ of values for each attribute $a \in A$. 
\end{definition}

At this stage, the definition of datasets is relatively unconstrained. In particular, there is no restriction on the types of the values used to describe each object of the dataset. As we will see later, most approaches to taxonomy construction from structured data however have limitations regarding the types of values they can take as input. In practice, it is also often assumed that all the values for a given attribute are of the same type. \\

The notion of concept is widely used in various areas, including description logics, formal concept analysis, ontologies, and machine learning, to represent a set of entities either present in the data or in a wider interpretation domain. Here, we adopt a definition inspired by formal concept analysis.

\begin{definition}[Concept]
A concept in a dataset $D=(O,A)$ is defined as a pair $(I,E)$. $I$, the intension, is a set of necessary and sufficient conditions for an object $o \in O$ to be considered a member of the concept. $E$, the extension, is defined as the set of all objects in $O$ that validate all conditions in $I$ and therefore belong to the concept. 
\end{definition}

In other words, a concept is considered a set of objects from the dataset (the interpretation domain) that is characterized by a set of conditions (the intension). Those conditions naturally relate to the values of the attributes that are used to describe each object in the dataset. We remain vague on the definition of those conditions at this point as it is another of the aspects that is specific to different approaches to taxonomy building from structured data. \\

While concept taxonomies generally include both the notion that a concept represents a set of objects (extension) and that they are associated with a given definition (intension), some approaches do not go as far as to make the intensional part of concepts explicit. We therefore here make use of a slightly different, weaker notion corresponding to groups of data objects defined purely extensionally (i.e. by enumeration). 

\begin{definition}[Cluster]
A cluster $C \subseteq O$ in a dataset $D=(O,A)$ is defined as a set of data objects from $O$ that are explicitly stated as being grouped together. 
\end{definition}

Clusters can be built based on various criteria, but those criteria are not expressed in the form of a set of conditions on the values of attributes of objects. \\

Finally, a taxonomy is more than a set of concepts: It organizes them in a hierarchy from more general to more specific. 

\begin{definition}[Concept subsumption]
A concept $A=(I_A, E_A)$ is said to subsume ($B \sqsubseteq A$) a concept $B=(I_B, E_B)$ if and only if $E_B \subseteq E_A$. 
\end{definition}

In other words, concept subsumption corresponds to the notion of a generalization relation between concepts. Following this definition, if a concept subsumes another, its intension will also be less constraining (with a lower number of or weaker conditions). The base notion of concept subsumption as defined through the inclusion of extensions naturally also applies to clusters. \\

\begin{definition}[Taxonomy]
In this work, a taxonomy is defined as a set of concepts, each associated with a label, and organized according to the subsumption relation.
\end{definition}

In other words, we consider a taxonomy to be a hierarchy of concepts, with every concept being given a name. 

\section{Related Work}
\label{sec:rw} 

Based on the definitions above, we consider the task of semiautomatic taxonomy building from structured data as one that takes as input a dataset and produces as output a structure as close as possible to a taxonomy. We consider three prominent families of approaches here, i.e. those based on formal concept analysis, on clustering, and on decision trees. Although these approaches have also been used for the bottom-up construction of taxonomies from text (see, for example,~\cite{cim1,cim2}), the corresponding methods do not apply directly to structured datasets as considered here. We therefore do not discuss those works here.

\subsection{Approaches based on formal concept analysis} 

Formal concept analysis (FCA)~\cite{FCA}, broadly summarized, consists in building concept hierarchies from formal contexts, which are representations of data objects with binary attributes. In other words, in formal concept analysis, the input dataset $D=(O,A)$ is such that any $o \in O$ is represented by a binary vector. 

The hierarchy built through formal concept analysis is formally a lattice in which a concept's intension is a binary vector, an itemset, indicating which attribute has to be true for a data object to belong to the extension. Those itemsets are said to be closed in the sense that all the elements of the extension of a concept have to contain the itemset in the intension (by definition), but also that no other attribute (item) should be shared by all elements of the extension in addition to the ones in the intension of the concept. In other words, if in a dataset $(O,A)$, we call $I'\subseteq O$ all objects that share all the items of the itemset (intension) $I$, and $E'\subseteq A$ all the attributes (item) shared by all the objects of $E$ (extension), then the concept $(I,E)$ is closed if $I''=I$ and $E''=E$.

Since it is building concept hierarchies from structured data, FCA has naturally been applied multiple times to support the semiautomatic creation of taxonomies. In~\cite{contento} for example (applied to the field of data licenses in~\cite{licensecontento}), it is used to create an initial hierarchy of concepts from the data, which is then manually pruned and labeled to form the core of an ontology. Many other works have made use of FCA to build taxonomies or parts of ontologies~\cite{taxoFCA,taxoFCA2}, or to align and merge them~\cite{fcamerging,fcamerging2}. 

From these works, we can see that the principles of FCA represent a strong foundation for semiautomatically building taxonomies from data. Indeed, apart from the labels of concepts, which would have to be entered manually, what is produced perfectly fits the definition of a concepts taxonomy as considered above. One of the key issues however is that all possible closed concepts are created, which often leads to very large and very complex hierarchies, from which it is difficult to figure out what is relevant and what can be discarded. In other words, editing a concept lattice created through FCA can potentially take more effort than manually building the taxonomy. 

Furthermore, FCA suffers from a number of technical limitations. First, it requires that objects in the dataset are described through binary attributes. Although categorical data can always be encoded and numerical data discretized, a certain level of precision can be lost in the process. In addition, those steps of encoding and discretization significantly increase the number of attributes in the data, and FCA is known to have issues in scaling both in number of objects and in number of attributes. As a result, FCA is really only applicable to relatively small datasets of low dimensionality. 

\subsection{Approaches based on clustering}

Clustering represents a family of machine learning techniques to automatically, in an unsupervised way, find groups (clusters) of data objects in a dataset, based on the values of their attributes~\cite{clustering}. Most clustering techniques are based on measuring the similarity of data objects and aim to create homogeneous clusters, that is, clusters of data objects similar to each other and dissimilar with those outside the clusters. 

The majority of clustering techniques (including popular ones such as K-Means and all its variants) are not suitable for building taxonomies since their output corresponds to a set of clusters (adhering to the definition given previously) that are not organized in a hierarchy. Despite this, some works such as~\cite{taxoClu} have applied these techniques to support the creation of taxonomies. In addition, methods such as agglomerative clustering~\cite{AC} (a.k.a., bottom-up hierarchical clustering) are a particular set of techniques that rely on the creation of a hierarchy of clusters in which every cluster is a set of data objects that is a subset of its parent cluster. Those hierarchies are therefore based on a relation that corresponds to the one of subsumption introduced earlier. 

As an example of an application of hierarchical clustering for taxonomy building, in~\cite{taxoAC}, the authors use this approach on a dataset of articles on the topic of design research, as a starting point for the creation of a taxonomy of design research issues. In~\cite{taxoAC2}, a simmilar process is used to build a taxonomy of IT subjects.

An obvious disadvantage of the use of (hierarchical) clustering for building taxonomies is that those methods produce clusters and not concepts, i.e., they do not produce an intensional definition for the groups of data objects they create. The clusters are formed on the basis of a notion of similarity that can be difficult to reconcile with providing clear definitions (necessary and sufficient conditions) for the set of objects they represent. 

In addition, hierarchical clustering can suffer a similar issue to FCA: It naturally produces a large number of clusters in a hierarchy that can therefore be very deep. Sorting relevant from irrelevant clusters, editing them to reflect useful categories in the domain, and labeling them can, therefore, require significant effort. 

However, on the technical side, those clustering techniques are less constrained than FCA: They can deal with a wider variety of attribute types and larger datasets. It is worth mentioning still that most techniques only work on numerical inputs (so nominal attributes need to be encoded), and remain slow and memory heavy on large datasets. 

\subsection{Approaches based on decision trees}

Decision trees are a popular form of machine learning models and are often used in classification and regression tasks. The main idea is to progressively divide the dataset into subsets based on basic conditions on the values of attributes in the dataset, in order to create subsets that are increasingly more homogeneous with respect to a given target variable (see, for example,~\cite{DTs}). Interestingly, nodes in a tree therefore correspond to the given definition of concepts, since they represent sets of data objects (extension) with an intensional definition (the condition attached to the node, in conjunction with all the conditions of the parent nodes). The relation on which the tree is built also corresponds to the one of subsumption, and adding labels to those nodes would achieve making it a concept taxonomy. 

A key issue with decision trees is that they are designed to be used in supervised machine learning tasks. This means that the construction of the decision tree, the successive splittings of the dataset, is driven by the goal of being able to predict the value of a given output variable. This was used, for example, in~\cite{DTTaxo} for the related task of concept induction. In this case, however, decision trees were used to automatically derive intensional definitions (in description logic) of concepts for which the label and extension were already known. We consider here a broader case where we are trying to build a taxonomy that is not tailored to a specific task or a specific objective. 

\subsection{Filling the gap}

Based on the review of the three categories of approaches described above, our goal is to take inspiration from them in order to build a method to create, with little effort from the human user, taxonomies that can be directly integrated into ontologies. In other words, the concepts in those taxonomies should be defined intentionally and divide the dataset hierarchically in a way that is driven by the data in input. 

To achieve that, we take an interactive and iterative approach: Instead of building an entire structure that needs to be heavily edited by the user in a second step, we take inspiration from decision trees in the way they progressively subdivide the dataset. Contrary to decision trees, however, first, this division is not driven by a specific task but by grouping data objects in homogeneous clusters. Importantly, the user can (and is encouraged to) intervene on the emerging taxonomy at any step to label the clusters created and provide an intensional definition for them, therefore making them concepts. These definitions, in the form of conditions on the values of attributes, have a direct effect on the extension of the concepts (which data objects are being grouped) and, therefore, on subsequent divisions. By enabling the user to carry out small editing actions at the time when the taxonomy is being built, we avoid creating a large, complex structure requiring a much more significant editing effort afterwards. 

\section{Interactive and Iterative Approach}

We expect the user of the process being considered here to be someone who is trying to build a taxonomy of a particular type of entities, so that the taxonomy is meaningful, useful, and well aligned with available data describing some of those entities. Also, while the data in question should be broadly understandable by this user, it might be too broad or complex to enable them to build such a taxonomy by manually exploring it. Further, while there might be multiple uses of such a taxonomy, the one we are aiming to support is its integration into an ontology, i.e. that it will form one component/one part of an ontology in which the type of entities described by the data is a core concept. 

To achieve this and illustrate the process of iterative and interactive clustering for taxonomy building from data, we take a simple but realistic example of a dataset of characters from comic books.\footnote{\url{https://github.com/rfordatascience/tidytuesday/blob/master/data/2018/2018-05-29/week9_comic_characters.csv}} This dataset takes the form of a table (a CSV file) where each line represents a comic book character, and each column a descriptor for those characters. Therefore, it naturally fits our definition of a dataset $D=(O,A)$ where $O$ is a set of data objects that each represents a character, and $A$ is a set of attributes. There are more than 20,000 characters described in this dataset ($|O|>20,000$) through 11 attributes ($|A|=11$). Those attributes include the publisher of the comic books where the characters appear (DC comics or Marvel), whether they are good or bad, their gender, hair and eye colors, date of first appearance, etc. Our objective here is therefore to build a taxonomy of comic book characters that is built from and aligned to those data, while having control over the process.

The idea of iterative and interactive clustering is that the dataset can be divided progressively, in a top-down manner, through a clustering method, and the results of those basic divisions be editable by the user at the time they are created. We therefore start with one concept, which will include as an extension the whole set of objects in the dataset (all the comic book characters), and with an empty intension.

From this step, the user can divide this initial concept into two parts using an automatic clustering method such as K-Means. What is obtained are two clusters that are based on grouping objects through their similarity. Inspecting each of the clusters, the user might notice that one contains most of the Marvel characters and the other most of the DC characters. Other characteristics are different between those two clusters, such as that one has more characters with public identities and the other has more with secret identities, or that the average year of first appearance of characters in one is earlier than in the other. 

Having considered the characteristics of the created cluster, the user can decide to give them a proper definition in the form of conditions on the values of their attributes. In our case, this would mean, for example, fixing the intention of the first cluster by adding the condition that the value of the publisher attribute is equal to ``Marvel'', therefore transforming the cluster created automatically into the concept of ``Marvel Character'' (which the user might decide to label it at this point). 

As the intension of a concept is being fixed, two things happen. First, the extension of the concept is recalculated so to correspond to the intension, hence here placing all and only Marvel characters in the newly defined concept. Second, the cluster present at the same level is automatically given an intension that corresponds to the complement of the one just created. In this case, there is only one other value for the attribute publisher: DC Comics. Therefore, the other cluster is transformed into a concept with one condition in its intension: $publisher=DC$.

At the next iteration, the user can choose one of the newly created concepts and repeat the process. Dividing in the same way the concept ``Marvel Character'', we might obtain a cluster where the majority of characters are ``Bad Characters'' (attribute $align$) and another where the majority are ``Good Characters''. Hence, the concept ``Marvel Villain'' can be created by fixing the value of $align$ to $bad$ from the first cluster. Here, there is another value for the attribute $align$: $changing$. Therefore, the counterpart to ``Marvel Villain'' is created, with a condition in its intension indicating that the value of the attribute $align$ is either $good$ or $changing$. This concept might be kept as is and named by the user, or further refined by restricting it to only one of the values (e.g., $good$, for the concept ``Marvel Hero''), leading to the automatic creation of a third concept, complement of the two others, with $changing$ as value for the $align$ attribute.

While the two examples above relate to attributes with nominal values, the same principle is used when the definition of a concept is based on a numeric attribute (such as the year of first appearance), with conditions indicating whether the value of the attribute is greater than or lower than a certain value. In this way it should be possible to distinguish for example ``first generation Marvel heroes'' from those of later generations. 

The pseudocode in Algorithm~\ref{pseudocode} summarizes the process informally illustrated above. In this code, the constructed hierarchy is represented by a root node with four elements: the set of objects that the node represents (the extension), the conditions defining the node (the intension), the label of the node, and the list of its descendants. Clusters are created with empty intensions, labels, and descendants. Creating a condition in line~10 corresponds to the user choosing an attribute $a$ from $A$ and a set of possible values for it if $a$ takes nominal values, or a treshold in the values if $a$ takes numeric values. 

\begin{algorithm}
\caption{Pseudocode of the process for taxonomy building from data}
\label{pseudocode}
\begin{algorithmic}[1]
\State $D=(O,A)$ \Comment{The dataset.}
\State $H\gets(O,\emptyset,\_,\emptyset)$ \Comment{The hierarchy}
\While{$not~done$}
\State $C\gets$ a leaf concepts chosen by the user from $H$
\State $Ac\gets$ the user selected action on $C$ 
\If{$Ac$ is $cut$}
     \State $C1,C2\gets$ results of clustering the extension of $C$
     \State Set $(C1, \emptyset, \emptyset)$ and $(C2, \emptyset, \emptyset)$ as descendants of $C$ in $H$
\ElsIf{$Ac$ is $define$}
    \State $c\gets$ condition created by the user for $C$
    \State Add $c$ to the intension of $C$
    \State Recalculate the extension of $C$
    \State Replace all not explicitly (by the user) defined siblings of $C$ by the complement of $C$ and its explicitly defined siblings
\ElsIf {$Ac$ is $label$}
    \State $l\gets$ user entered label fo $C$
    \State Set the label of $C$ to $l$
\EndIf
\EndWhile
\end{algorithmic}
\end{algorithm}

As visible here, this process is designed to enable quickly identifying candidate concepts from potentially large data, but leaves the user in control of the definition of those concepts at every step, therefore minimizing the editing effort required, and ensuring that the resulting taxonomy is aligned with both the knowledge of the user and with the data from which it was built. 

\section{Implementation}

\begin{figure}[p]
    \centering
    \rotatebox[origin=c]{270}{\includegraphics[width=20cm]{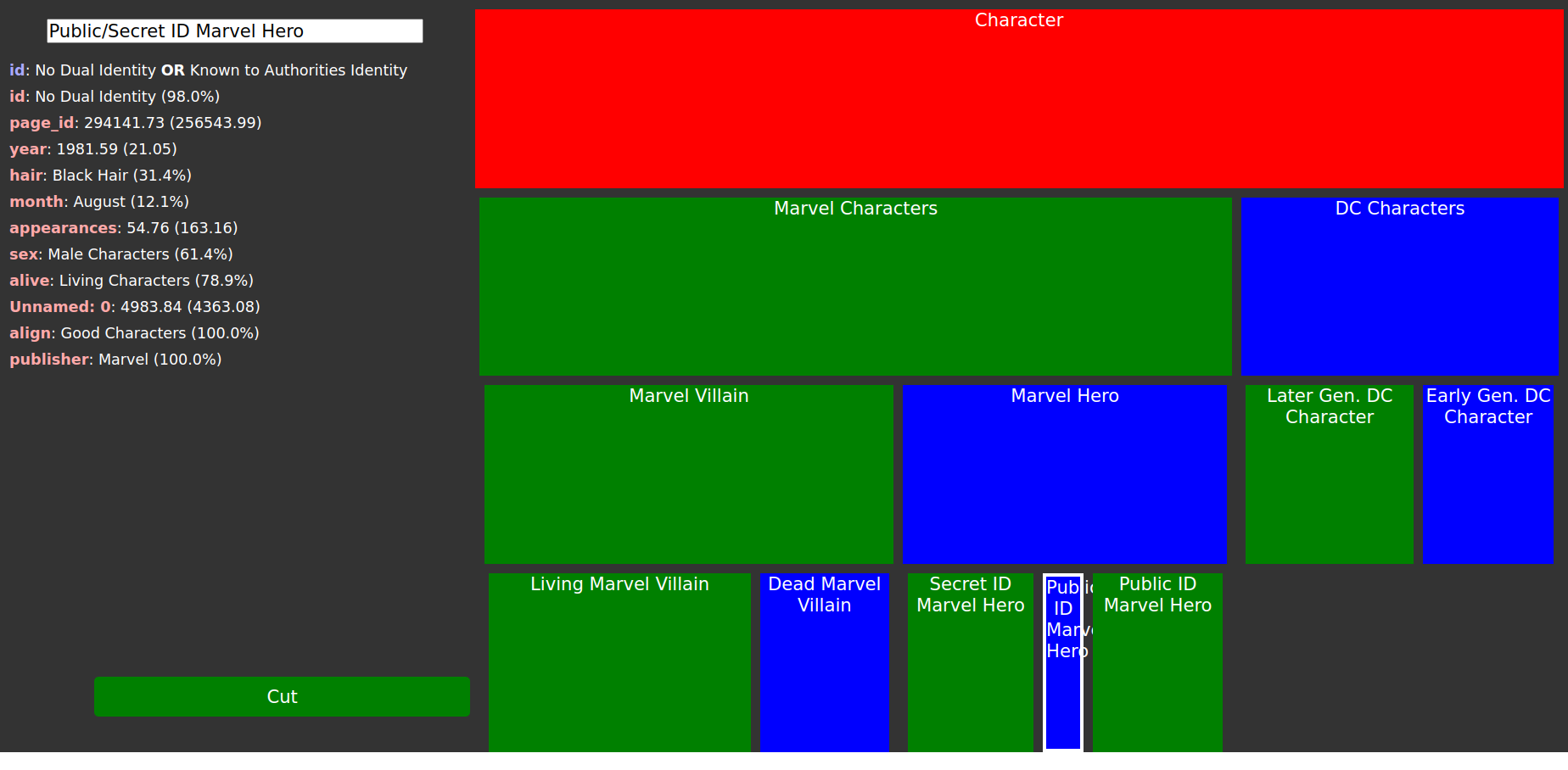}}
    \caption{Interface of the TaBIIC tool applied to the comic book character dataset. The dataset has first been divided by publisher (DC or Marvel). We then divided DC Characters based on the date of first appearance. Marvel characters were divided between heroes and villains, with heroes being further divided based on whether they have a public, secret or mixed public/secret identity. Finally, Marvel villains were divided according to whether or not they are alive in the most recent comic book series.}
    \label{fig:ss}
\end{figure}

A proof of concept implementation of the process described above is available online\footnote{\url{https://github.com/mdaquin/TaBIIC}} (see screenshot Figure~\ref{fig:ss}). It takes the form of a tool called TaBIIC (Taxonomy Building through Iterative and Interactive Clustering), which allows a user to load a dataset in the form of a CSV file, select the variables to be considered in the construction of the taxonomy, divide the dataset and already formed concepts through clustering, provide intensional definitions for those concepts and label them. 

In this prototype, a number of choices have been made to simplify the role of the user and decrease the amount of effort required. This includes pre-processing the data so that rows with missing values are removed, numeric data are processed, and identifiers are automatically excluded from the variables used in the taxonomy building process. This also includes providing a single pre-configured approach for dividing concepts into clusters. Here, we use the K-Means algorithm with the parameter $k$ (number of clusters to create) set to 2. Although this algorithm might not always give perfect results, the fact that the user always has the option to correct those results by specifying the intensional definition of concepts makes these imperfections less impactful.

Among other useful features, TaBIIC includes a panel on the left of the interface to display details of the values of attributes for any selected cluster/concept. This includes, first, any intensional definition not inherited from parent concepts (in Figure~\ref{fig:ss}, that the attribute $id$ takes the values $No~Dual~Identity$ or $Known~To~Authorities~Identity$). This is followed by statistics for each attribute, that is, average and standard deviation for numeric attributes or the most frequent value and its frequency for nominal attributes. The attributes here are ranked based on the difference in those statistics with sibling concepts/clusters. In other words, this puts first attributes that seem to best characterize the objects in the concept or cluster, as opposed to those outside of it. 

\begin{figure}
    \centering
    \includegraphics[width=\linewidth]{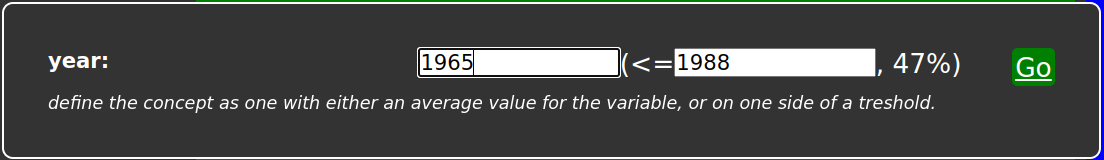}
    \caption{Interface for entering a condition for a numeric variable. In the example here, we are looking to define the concept of DC characters having an average year of first appearance of 1965, which corresponds to the DC characters in the dataset who appeared first before 1988. It is also specified in the interface that those represent 47\% of the DC characters.}
    \label{fig:numdef}
\end{figure}

Editing the intensional definitions of concepts is done through first selecting, in the panel described above, the attribute on which the user wishes to set a condition. If the attribute is nominal, the list of possible values for it can be entered by separating each value with the keyword $OR$. The input field is pre-filled with the most frequent value in the concept/cluster. If the variable is numeric, the user can enter either the target average value in the concept or the threshold below or above which an object will be considered part of the concept (with the dependency between those two values being automatically calculated, see Figure~\ref{fig:numdef} for an example). The input field for the average is pre-filled with the average value of the attribute in the cluster/concept currently being edited. 

From a technical point of view, the backend of the tool is written in Python, using the libraries scikit-learn for K-Means and pandas for data manipulation, and flask as a web framework. The interface is written in HTML/CSS/Javascript. 

\section{Evaluation and Discussion}

As described previously, we expect the interactive and iterative approach implemented in the TaBIIC tool to reduce the effort required to come up with a taxonomy that is valid, aligned with the data and that can easily be integrated into an ontology. We also expect it to be applicable efficiently to a variety of different datasets. To provide initial tests of these assumptions, we applied the tool on three datasets of varied sizes and richness. We discuss the results in the following. \\

The first dataset on which we tested TaBIIC is the toy IRIS dataset\footnote{\url{https://archive.ics.uci.edu/ml/datasets/iris}} often used as an example to illustrate machine learning methods. This dataset includes the description of 150 iris flowers according to the length and width of their petals and sepals. Each flower is associated with one of three species of iris (50 flowers per species). 

\begin{figure}
    \centering
    \includegraphics[width=0.50\linewidth]{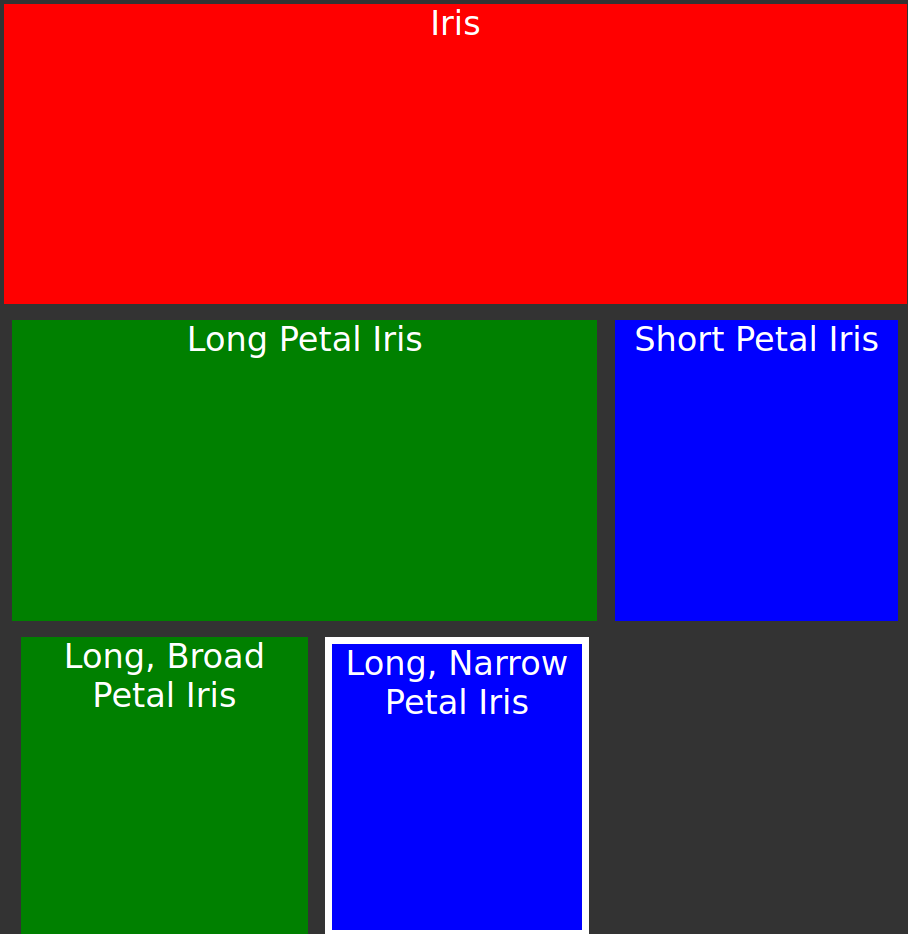}
    \includegraphics[width=0.49\linewidth]{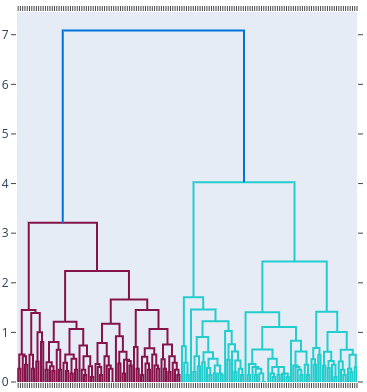}
    \caption{TaBIIC-created taxonomy (left) and hierarchy from agglomerative clustering (right) for the iris dataset.}
    \label{fig:iris}
\end{figure}

The dataset being very simple, building a taxonomy for it with TaBIIC did not raise much challenge and computation time was in particular negligible at every step. In applying TaBIIC, we naturally excluded the variable corresponding to the species. The resulting taxonomy is shown in Figure~\ref{fig:iris}, and consisted of first separating (based on the most differentiated attribute, and using the suggested value) the flowers according to the length of their petals, and then further dividing the ``long petal irises'' based on the width of their petal. Interestingly, the leaf nodes of the created taxonomy correspond almost exactly to the species of iris as classified in the dataset. 

Considering that this case is far from challenging, we can compare it with another approach, which could be too cumbersome to apply on datasets with more dimensions. In particular, we used bottom-up hierarchical clustering (agglomerative clustering) to create a hierarchy based on the similarity between the values of the four variables that describe each iris flower\footnote{We do not apply FCA here since, even on such a small scale example, the added complexity from having to transform attributes into ones with binary values, therefore multiplying their numbers, would require significant computation and necessarily lead to results that are cumbersome to exploit.}. The result is shown in Figure~\ref{fig:iris} next to the TaBIIC-created taxonomy. While it is obvious that the result from hierarchical clustering is much less interpretable than the one obtained with TaBIIC, it is not a fair comparison since its construction would generally be followed by a level selection step, reducing it to the top levels of the hierarchy at a depth of the user's choosing. Depending on the configuration of the algorithm, cutting it at level 2 in this case could lead to similar results as provided by TaBIIC. However, choosing this level and understanding the content and possible definition of the clusters produced would certainly require more effort even in this simple case. \\

The second dataset on which we tested is the comic book character dataset that we have already described. While it contains more than 20,000 rows, once rows with missing values are removed, only about 9,500 remain. This dataset is more realistic, as it is larger in scale, but also since it includes more attributes, with a mix of nominal and numeric values. Because of this, there is more user choice involved in the construction of the taxonomy. As shown earlier, the first division naturally separates characters from Marvel comics from characters from DC comics. Following this, whether characters are good or bad, have a secret identity, or their gender tend to appear as highly divisive attributes, but the user is free to construct the taxonomy on whichever criteria they choose based on the information provided. 

Since the dataset is larger, the computation time increases significantly, although K-means clustering in the worst case (the first division) did not take more than a few seconds. The hierarchy created by agglomerative clustering cannot be shown here as it is extremely large. It took several minutes to display on a recent laptop. Also, it is clear, considering the choices made during the construction of the taxonomy with TaBIIC that, first, choosing the right level at which to cut this hierarchy would be far than trivial, and second, that the results might require extensive adjustments to match the need of the user. \\

The last dataset\footnote{\url{https://www.kaggle.com/datasets/antoinekrajnc/soccer-players-statistics}} we tested contains many descriptors (more than 50) of soccer players (including mostly statistics and ratings for different aspects of the game). It is large (over 17,000 rows, even once rows with missing values are removed), and most of the attributes are numeric. It is a much more challenging dataset in the sense that it can be difficult to understand which attribute to select when creating a condition. This is not helped by the fact that many of those attributes are highly correlated, which means that choosing one over the other is unlikely to make a significant difference. 

To start with the technical aspects, despite working on a more challenging dataset, the tool remains usable with the soccer dataset. Dividing even the root node takes about half a minute on a recent laptop, and other operations, while slower, remain within acceptable times. It is worth mentioning that we did not manage to obtain a hierarchy from agglomerative clustering on this dataset. 

Despite being challenging and therefore requiring more effort, our tests show that it is possible, sometime after multiple attempts through trial and error, to structure the dataset with TaBIIC to differentiate, for example, highly rated from lower rated players, older from younger players, or players stronger in defense from those stronger in attack, and to combine those criteria to create a reasonable taxonomy. An interesting aspect here is that this process is not only helpful in creating the taxonomy, but also provides a way for the user to explore the dataset and gain an understanding of its content. \\

Based on the results of those tests, we believe that the approach presented here is not only feasible but represents an efficient (in terms of computing time and user time) approach to building taxonomies from data, which can be applied on a variety of datasets. As mentioned earlier in this paper, the objective was also that the taxonomy created could easily be integrated into an ontology. The definition we relied on for concepts is indeed consistent with the one used in ontology formalisms such as OWL, especially those based on description logics. In addition, the types of conditions used for nominal attributes can naturally be represented in those formalisms through value-based restrictions and concept unions. The conditions based on numeric values require the ability to apply restrictions on concrete domains, which is feasible according to the specification of OWL, and is supported by a number of description logic reasoners. Finally, it could be remarked that the taxonomies created by TaBIIC are trees rather than more general hierarchies and that, by construction, the set of direct descendants of a given concept represents a partition of that concept. This can be represented through disjointness axioms in OWL.

\section{Conclusion}

In this paper, we presented an approach to semiautomatically build taxonomies from data, using an iterative and interactive process where the user can control and edit the definitions of concepts at every step. Through several tests, we showed that the approach is conceptually and technically feasible and, through the proof-of-concept implementation presented, provides an efficient tool to structure and organize data into a taxonomy. We further discussed how that taxonomy can be straightforwardly integrated as a component of an ontology. In addition, it is worth mentioning that besides the practical objective of building a taxonomy, this tool provides an interesting way to explore a dataset in a potentially unknown domain and gain an understanding of its inherent structure. 

The tool presented can at this stage only be considered a proof of concept, and following the observations presented above, a key aspect of future work relates to strengthening this implementation so as to enable conducting more extensive user studies on the approach. Some aspects could be straightforwardly improved, including the ability to select different clustering methods and the actual export of the created taxonomy in the OWL format. Related to the former, an aspect that can be interesting to study is how the clustering technique used, as well as its parameters, could affect the quality of the end result (the taxonomy created). This kind of question can also be extended to the quality of the data. Even if the user can control the construction of the taxonomy at every step, it is expected for noise in the data to still have an effect, especially considering that we have implicitly assumed here that a taxonomy can be reliably created from a snapshot of relevant data at a given time. Maintaining the alignment between the created taxonomy and evolving data is therefore another potentially interesting research direction following this work (for example, in the spirit of~\cite{evolva}).

\bibliographystyle{plain}
\bibliography{8611}
\end{document}